\documentclass[lettersize,journal]{IEEEtran}
\usepackage{amsmath,amsfonts}
\usepackage{algorithm}
\usepackage{array}
\usepackage[caption=false,font=normalsize,labelfont=sf,textfont=sf]{subfig}
\usepackage{textcomp}
\usepackage{stfloats}
\usepackage{url}
\usepackage{verbatim}
\usepackage{graphicx}
\usepackage{cite}
\usepackage{cite}
\usepackage{booktabs}
\usepackage{multirow}
\usepackage{graphicx} 
\usepackage{hyperref}
\usepackage{makecell}
\usepackage{mathtools}
\usepackage{algpseudocode} 

\usepackage{color}
\usepackage[table]{xcolor} 

\begin{document}

\title{Vision-Language Model Purified Semi-Supervised Semantic Segmentation for Remote Sensing Images}

\author{\IEEEauthorblockN{Shanwen Wang, Xin Sun,~\IEEEmembership{Senior member,~IEEE}, Danfeng Hong,~\IEEEmembership{Senior member,~IEEE}, Fei Zhou
}
\thanks{This work is supported by the Science and Technology Development Fund - International Collaborative Research, Macao SAR (0001/2025/AIJ), Science and Technology Development Fund, Macao SAR - Ministry of Science and Technology: National Key R\&D Program of China (0007/2025/AMJ, 2025YFE0202900), and Science and Technology Development Fund, Macao SAR - Basic Research (0006/2024/RIA1)} 
\thanks{S. Wang and X. Sun are with Faculty of Data Science, City University of Macau, 999078, SAR Macao, China.}
\thanks{D. Hong is with School of Automation, Southeast University, Nanjing, 211189, China.}
\thanks{F. Zhou is with College of Oceanography and Space Informatics, China University of Petroleum (East China), Qingdao, 266580, China}
}

\markboth{Journal of \LaTeX\ Class Files,~Vol.~14, No.~8, August~2021}%
{Shell \MakeLowercase{\textit{et al.}}: A Sample Article Using IEEEtran.cls for IEEE Journals}

\maketitle

\begin{abstract}
The semi-supervised semantic segmentation (S4) can learn rich visual knowledge from low-cost unlabeled images. However, traditional S4 architectures all face the challenge of low-quality pseudo-labels, especially for the teacher-student framework. We propose a novel SemiEarth model that introduces vision-language models (VLMs) to address the S4 issues for the remote sensing (RS) domain. Specifically, we invent a VLM pseudo-label‌ purifying (VLM-PP) structure to purify the teacher network's pseudo-labels, achieving substantial improvements. Especially in multi-class boundary regions of RS images, the VLM-PP module can significantly improve the quality of pseudo-labels generated by the teacher, thereby correctly guiding the student model's learning. Moreover, since VLM-PP equips VLMs with open-world capabilities and is independent of the S4 architecture, it can correct mispredicted categories in low-confidence pseudo-labels whenever a discrepancy arises between its prediction and the pseudo-label. We conducted extensive experiments on multiple RS datasets, which demonstrate that our SemiEarth achieves SOTA performance. More importantly, unlike previous SOTA RS S4 methods, our model not only achieves excellent performance but also offers good interpretability. The code is released at \href{https://github.com/wangshanwen001/SemiEarth}{https://github.com/wangshanwen001/SemiEarth}.
\end{abstract}


\section{Introduction}
\IEEEPARstart{R}{mote} sensing (RS) image semantic segmentation is a core technology for understanding surface cover and monitoring environmental changes \cite{zhao2025pamsnet,hong2026hyperspectral,luo2025domain}, but it heavily relies on large-scale annotated data \cite{sun2025rsprotosemiseg, hong2024spectralgpt}. At the same time, enormous unannotated RS images are acquired from satellites and drones, which are an unexplored treasure to understand the Earth. Semi-supervised semantic segmentation (S4) can significantly reduce annotation requirements, which is an effective technique for such data. In RS scenarios, S4 specifically enhances the generalization ability in practical applications such as farmland monitoring and disaster assessment with limited labeled data \cite{hong2023cross, zi2025semi, feng2025improving}. So far, S4 research primarily focuses on efficient pseudo-label generation \cite{10771804}, consistency regularization \cite{11062866}, and self-training strategies \cite{wang2022self} to mine the potential information of unlabeled data. However, traditional S4 architecture ‌cannot address the challenges posed by low-quality pseudo-labels‌. Specifically, teacher-student architectures suffer from low-quality pseudo-labels generated by the teacher network, which can mislead the student model \cite{11062866}. Self-training strategies \cite{11145106} and consistency regularization methods like FixMatch \cite{sohn2020fixmatch}, which rely on a single network, also produce unreliable pseudo-labels. The pseudo-labels generated by either a teacher or the model itself are the key to model training. However, these incorrect pseudo-labels often lead the model to unstable and erroneous results \cite{yang2025unimatch, ma2025dual}. Even though researchers recently focused on solving this problem, they still struggle with some drawbacks of common S4 architectures. For example, several works attempt to filter pseudo-labels based on confidence and uncertainty \cite{11062866, huang2023semi, lu2025uncertainty}, but the selected samples still contain a large amount of noise. It is noteworthy that low-confidence regions filtered out by confidence and uncertainty are located in ambiguous multi-class boundaries. And it will make the model ineffective if we exclude these pixels from training. Moreover, incorrect pseudo-labels and training errors will be amplified during training, especially in multi-class boundary areas that require fine segmentation.

\begin{figure}[!t]
\centering
\includegraphics[width=3.5in]{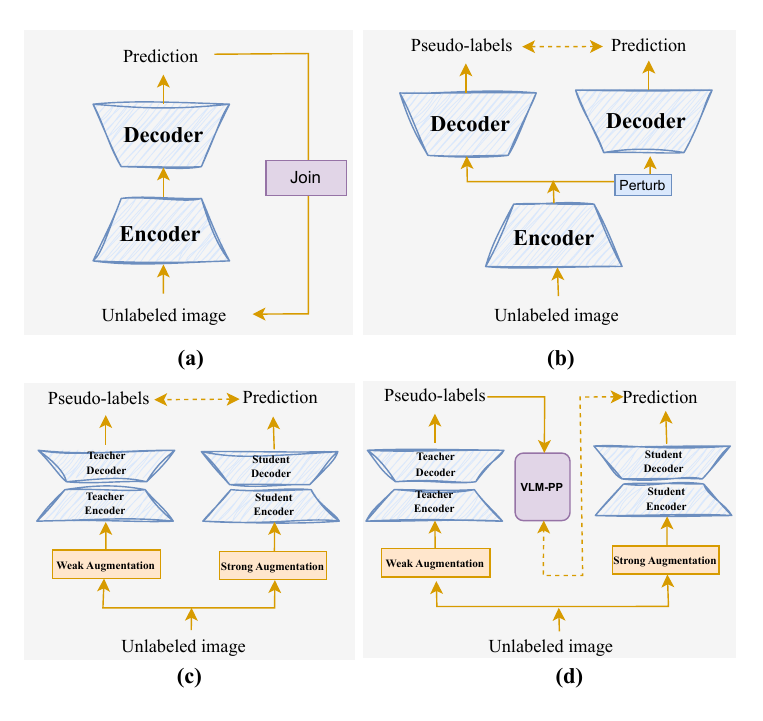}
\vspace{-0.7cm}
\caption{Structure comparison of S4 frameworks: (a) Self-training semi-supervised models, (b)-(c) Consistency regularization models, (d) Our SemiEarth.}
\label{fig1}
\vspace{-0.5cm}
\end{figure}

To address these issues, we innovatively introduce visual language models (VLMs) into the RS S4 domain, proposing a novel semi-supervised RS model called SemiEarth. SemiEarth abandons various complex mechanisms, such as multiple perturbations and various weighting schemes \cite{10771804}. Instead, it introduces VLMs to purify low-quality pseudo-labels. The schematic structures of SemiEarth and mainstream S4 methods are shown in Fig. \ref{fig1}. Unlike traditional S4 architectures that focus on perturbation and consistency, SemiEarth establishes a novel scheme that integrates visual-linguistic features. We introduce the VLM pseudo-label purifying structure (VLM-PP) for SemiEarth to purify low-quality pseudo-labels generated by the teacher model. Rather than simply discarding low-confidence pseudo-labels, VLM-PP seeks to refine and enhance the quality of those with extremely low confidence. More importantly, VLM-PP can rectify errors when the teacher model generates low-quality pseudo-labels with misclassifications. The main contributions of this paper are as follows:

\begin{enumerate}
\item We propose a novel semi-supervised model, SemiEarth, which for the first time introduces VLMs into RS S4 domain to address the challenge of low-quality pseudo-labels generated by teacher models.
\item We propose a VLM-PP module to purify low-quality pseudo-labels from the teacher, significantly enhancing pseudo-label quality and effectively preventing the student from being misled.
\item ‌We propose a rectification mechanism for false pseudo-labels in the RS S4 domain, ‌leveraging the VLM's independent judgment of low-quality pseudo-labels outside the common RS S4 framework.
\item Extensive experiments on RS datasets demonstrate that our model significantly outperforms traditional SOTA methods and offers very good interpretability.
\end{enumerate}

The rest of this article is organized as follows. Section \ref{RW} briefly introduces the related work. In Section \ref{METHODS}, our model, SemiEarth, is proposed and discussed. Section \ref{Experiments} shows the experimental results and compares them with SOTA methods. Section \ref{Conclusion} concludes this article.

\section{Related work}
\label{RW}
In recent years, with the maturity of deep learning, an increasing number of S4 models and VLMs have been applied in the field of RS. This section systematically reviews recent advances in S4 models and VLMs for RS, providing the necessary background for our proposed model.

\subsection{Semi-Supervised Semantic Segmentation}
The advent of deep learning has revolutionized the RS domain, with fully supervised networks like FCN \cite{damalla2025transrefine}, SegNet \cite{weng2020water}, and U-Net \cite{wu2022uiu} achieving remarkable progress in semantic segmentation. These models have promising segmentation capabilities and stability, but their success hinges on large-scale, high-quality labeled datasets, which are costly and labor-intensive to acquire. In contrast, RS S4 models aim to leverage abundant unlabeled data instead of well-labeled data, thereby reducing dependence on expensive pixel-level annotations \cite{han2025difference}. The recent RS S4 models primarily focus on two aspects of innovation. One is to address the inherent limitations of the S4 algorithms themselves \cite{10771804, li2021semisupervised}. For example, previous studies by Huang\cite{huang2024decouple}, Lu\cite{lu2025uncertainty}, and Chen\cite{chen2025tse} have tried to solve the problems of low quality of pseudo-labels and the inherent distribution mismatch between data. However, they targeted S4 structure improvement and mismatched the inherent characteristics of RS images, such as multi-scale information and the complex features of multi-class boundary regions. The other is to target the unique characteristics of RS images. For instance, Ni \cite{11129886}, Wang \cite{11062866}, and Xin et al. \cite{xin2024confidence} focused on addressing multi-scale features and high inter-class similarity specific to RS images. Specifically, Ni \cite{11129886} and Wang \cite{11062866} addressed the multi-scale challenges by leveraging label space for contextual label readjustment and employing multi-scale uncertainty consistency, respectively. Facing the high inter-class similarity in RS images, Xin \cite{xin2024confidence} and Wang \cite{11062866} adopted contrastive learning and cross-teacher-student attention networks. However, they designed quite complex structures and made limited performance improvement. For example, the method by Wang et al. \cite{11062866} only achieved a marginal improvement of about 1\% to 2\% in mIoU compared to the previous SOTAs. In addition, some RS S4 methods, including MCMCNet \cite {gao2024mcmcnet} and SemiRoadExNet \cite{chen2023semiroadexnet}, pay attention to specific categories of RS without generalizability.

Overall, these RS S4 methods in the RS field are constrained by traditional architectures, failing to accurately remove errors in pseudo-labels‌, especially for multi-class boundary areas requiring fine segmentation in RS images. Particularly, pseudo-labels pay significant role in the training procedure. This study introduces a pseudo-label purification module, VLM-PP, which is independent of the S4 architecture. It can effectively purify out low-quality pseudo-labels and rectify the teacher's errors, significantly improving the performance of the student.

\subsection{Vision-Language Models for Remote Sensing}

The existing deep learning methods in the RS field have primarily focused on visual processing while neglecting semantic understanding \cite{zhan2025skyeyegpt, dong2024changeclip, lang2024toward}. For example, visual models may misclassify building roof as highways when their visual features of pixels are similar. The main reason is that the model lacks common-sense knowledge that a highway cannot be on a building's roof. VLMs can jointly reason about images and their textual descriptions, thereby gaining a deep understanding of semantic relationships \cite{hu2025rsgpt, wang2024skyscript, zhu2025skysense}.  

Specifically, VLMs provide an opportunity to explore the integration of general and expert knowledge into visual analysis tasks for RS data \cite{li2024vision}. For instance, VLMs are aware that ships are likely to be located on water rather than on land \cite{liu2024rotated}. Therefore, VLM-based segmentation models often avoid misidentifying categories on the ground as a ship, demonstrating potential improvement for RS analysis. Currently, some studies have explored the application of VLMs in various RS analysis tasks, including RS image captioning (RSIC) \cite{lin2025rs,yang2022meta,zia2022transforming}, text-based RS image retrieval (RS-TBIR) \cite{xiong2022interpretable, bai2025textir}, RS visual question answering (RS-VQA) \cite{chappuis2025evaluating,wang2024earthvqa,chappuis2022prompt}, referring RS image segmentation (RRSIS) \cite{chen2025rsrefseg,liu2024rotated}, and open-vocabulary RS segmentation (OVRS)\cite{cao2024open,ye2024towards,li2025exploring}. 

However, the aforementioned applications of VLMs in the RS field are based on large-scale labeled dataset training. Meanwhile, some pioneer works, such as the SegEarth-OV\cite{li2025segearth} and RSCLIP\cite{wangrsclip} models, have explored the zero-shot task with VLMs in RS. While they have made important contributions, the improvement of performance is limited due to the inherent constraints of the CLIP model, which stems from its pre-training on image-text pairs. The purpose of this study is to integrate the VLMs with the RS S4 framework to fully utilize the unlabeled RS images. We aim to achieve excellent performance using only a small amount of labeled and a large amount of unlabeled RS image data for training. In addition, the proposed SemiEarth does not train the VLM model itself but only uses it for the purification of pseudo-labels during inference.

\begin{figure*}[!t]
\centering
\includegraphics[width=7in]{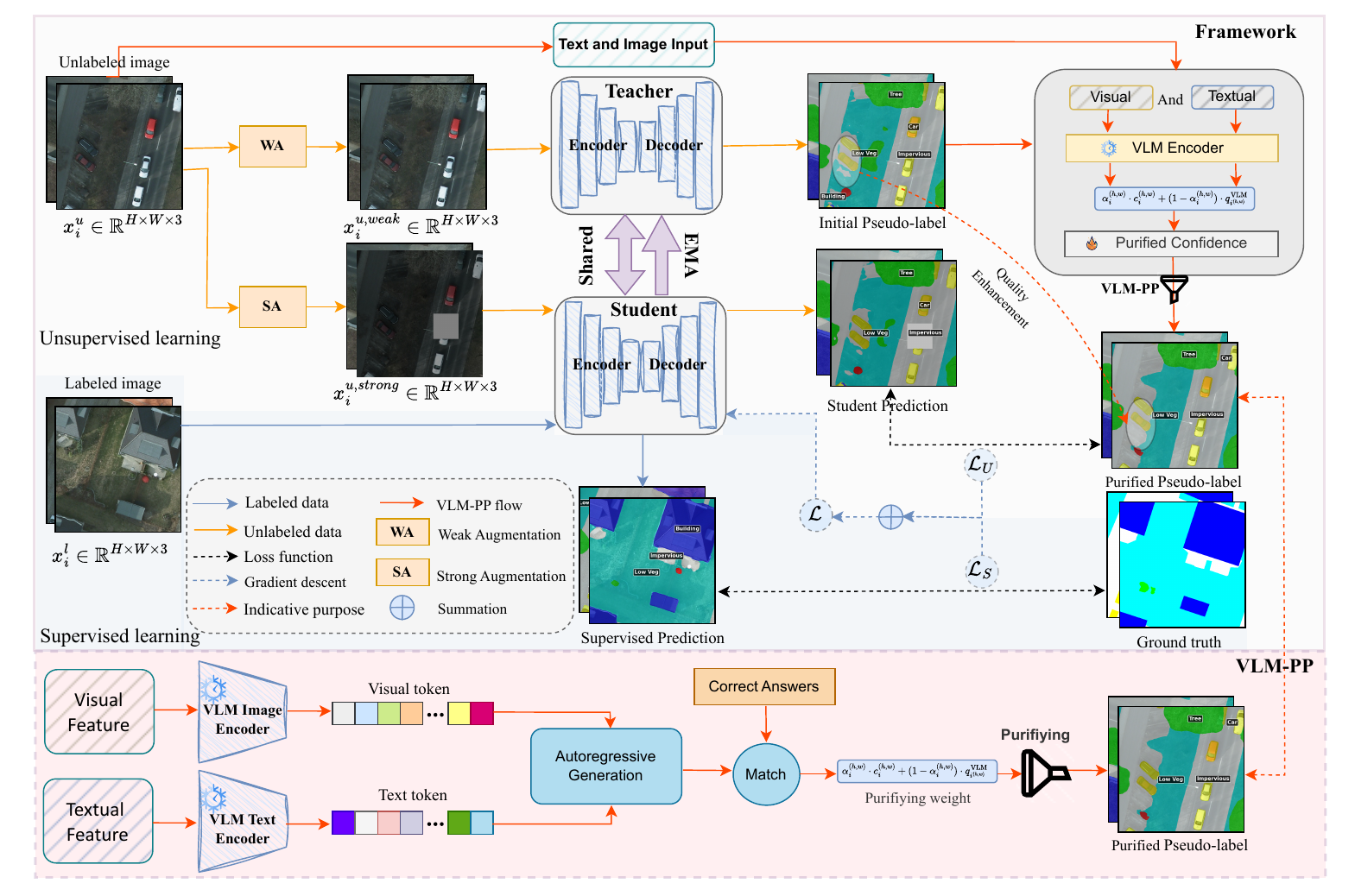}\vspace{-0.3cm}
\caption{The overall structure of our VLM-purified RS S4, i.e., SemiEarth, consists of unsupervised learning with unlabeled data and supervised learning with labeled data. The pseudo-labels generated by the teacher are not directly provided to the student but first purified through the VLM-PP. Blue lines indicate labeled data flow, yellow lines indicate unlabeled, red lines indicate VLM-PP flow, blue dashed line indicates gradient descent and backpropagation, and the red dashed lines serve as an indicative purpose.}
\label{framework}
\vspace{-0.3cm}
\end{figure*}

\section{Method}
\label{METHODS}
This section is organized as follows: Section \ref{model} describes our main framework, Section \ref{motivation} introduces the motivation ‌of the VLM-PP, and Section \ref{VLM-PP} describes the core principles ‌of the VLM-PP.

\subsection{Preliminaries and Main Framework}
\label{model}
Semi-supervised learning can train a model with a small amount of labeled data for supervised learning and a large amount of unlabeled data for unsupervised learning. The proposed SemiEarth is a specific semi-supervised learning model as shown in Fig. \ref{framework}. Specifically, we use light gray, light blue, and light beige regions represent unsupervised learning, supervised learning, and VLM-PP structure. SemiEarth takes both labeled and unlabeled images as training data for supervised and unsupervised learning, respectively. The VLM-PP is the method to combine and improve these two learning procedures. We first define $ D^L=\{(x_i^l,y_i)\}_{i=1}^{N_L} $ as labeled data and $ D^U=\{(x_i^u)\}_{i=1}^{N_U} $ as unlabeled. Here ${x_i^l}\in \mathbb{R}^{H\times W \times 3}$ denotes labeled image, ${y_i}\in \mathbb{R}^{H\times W \times K}$ is the ground truth, while ${x_i^u}\in \mathbb{R}^{H\times W \times 3}$ denotes unlabeled image. $N_L$ and $N_U$ are the amount of labeled and unlabeled images, where $N_L << N_U$. $H$ and $W$ specify the height and width of the image. As shown in  Fig. \ref{framework}, the supervised loss $\mathcal L_S$ constrains the supervised predictions of the student on labeled data and ground truth. Unsupervised loss $\mathcal L_U$ constrains the consistency between students' predictions on unlabeled data and pseudo-labels. The overall loss function $\mathcal L$ summarizes supervised $\mathcal L_S$  and unsupervised $\mathcal L_U$ losses as follows:
\begin{equation}
\label{deqn_ex1a}
\mathcal L=\mathcal L_S+\mathcal L_U =\frac{1}{N_L}{\sum_{i=0}^{N_L}\mathcal L_{CE}(p_i^l,y_i)}+ \frac{1}{N_U}{\sum_{i=0}^{N_U}\mathcal L_{CE}(p_i^{u,s},y_i^{u,t})},
\end{equation}
\noindent where $p_i^l$ is the prediction of labeled image $x_i^l$, $\mathcal L_{CE}$ is cross-entropy loss, $p^{u,s}_i$ is student predictions for unlabeled data $x_i^u$, and $y^{u,t}_i$ is the pseudo-label. 

Similar to most RS S4 architectures, SemiEarth emphasizes unsupervised learning. As shown in the light gray region of Fig. \ref{framework}, SemiEarth adopts basic teacher-student architecture for unsupervised learning. Specifically, the unlabeled data is first ‌applied with‌ weak and strong augmentations, respectively. Weak augmentation provides the teacher with stable features to generate reliable pseudo-labels, while strong augmentation enables the student to learn from a broader range of variations. Parameters of the student are updated under the guidance of pseudo-labels generated by the teacher. And the weights $\theta^t$ of the teacher model are updated via an exponential moving average (EMA) of the student's weights $\theta^s$. Specifically, the teacher's weight $\theta_t^t$ is updated at step $t$ as $\theta_t^t= \alpha\theta_{t-1}^t + (1-\alpha)\theta_{t}^s$, where $\alpha$ is the EMA decay. Therefore, the teacher gradually integrates information from different training steps through the EMA mechanism. This yields smooth and robust predictions, which then provide the student model with higher-quality pseudo-labels. However, due to the numerous categories and complexity of RS images, the pseudo-labels generated by the teacher often suffer from low quality and unreliability. Therefore, we purify them through a novel VLM-PP structure before providing pseudo-labels to student.

\subsection{‌Motivation for VLM-PP}
\label{motivation}
In this section, we state the motivation and justification for proposing the VLM-PP for the SemiEarth model.‌ The core challenge of RS S4 stems from the epistemic uncertainty inherent in pseudo-label generation. Unlike supervised learning, where ground truth provides unambiguous supervisory signals, pseudo-labels are constrained by the knowledge from the teacher. However, the correctness of the predictions from the teacher cannot be guaranteed. The teacher model's erroneous predictions directly provide students with incorrect pseudo-labels for learning, and these errors systematically propagate and amplify through iterative training. This creates a self-reinforcing loop where mistakes become increasingly entrenched rather than corrected. Existing approaches attempt to mitigate this issue through confidence-based thresholding, only retaining pseudo-labels with a predefined confidence value. However, it is noteworthy that low-confidence regions are often located in ambiguous multi-class boundaries or rare classes. And it will make the model ineffective if we exclude these pixels from training. More important, existing methods can not rectify the error pseudo-labels, i.e., the traditional S4 structure cannot transcend the knowledge boundaries of its own learned feature. Therefore, we introduce VLM-PP as external semantic knowledge to purify pseudo-labels.

\begin{figure}[t]
\centering
\includegraphics[width=3.5in]{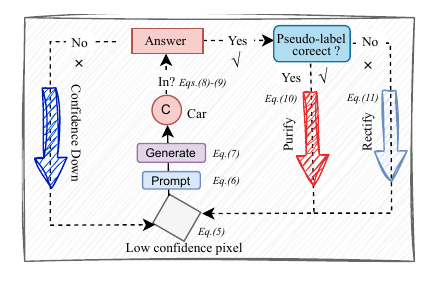}
\caption{The core logic of VLM purifying low-quality pseudo-labels from teachers.}
\label{VLM-PP_Fig}
\vspace{-0.3cm}
\end{figure}

We formulate the proposed purification procedure for logical reasoning and justification. First, we generate key pseudo-labels and confidence scores as follows:
\begin{equation}
\label{pixel}
p^{u,t(h,w,k)}_i=P(pixel_{(h,w)}= k|x^{u,weak}_i),
\end{equation}
\begin{equation}
\label{pseudo-labels}
y^{u,t(h,w)}_i=\mathop{\arg\max}_{k \in [1,K]}(p^{u,t(h,w,k)}_i),
\end{equation}
\begin{equation}
\label{confidence}
c^{(h,w)}_i=\mathop{\max}_{k \in [1,K]}(p^{u,t(h,w,k)}_i).
\end{equation}
Here, ${x_i^{u,weak}}$ represents the weakly augmented unlabeled data, $(h,w)$ represents pixel coordinates, $P$ is probability function, $k \in[1,K]$ is the class index, and $p^{u,t(h,w,k)}_i$ denotes the probability that the pixel at position $(h,w)$ belonging to class $k$. For each pixel, the class with the highest probability is the pseudo-label $y^{u,t(h,w)}_i$. The maximum probability value $c^{(h,w)}_i$ is the confidence. The closer of $c^{(h,w)}_i$ to 1, the more certain the model is. Conversely, the closer to $1/K$, the more uncertain the model is, which means almost random. We define the low-confidence region as follows.
\begin{equation}
\label{low-confidence regions}
\mathcal{R}_{\text{low}} = \{(h,w) \mid c_i^{(h,w)} < \tau_{\text{conf}}\},
\end{equation}
where $ \tau_{\text{conf}}$ is the VLM-PP confidence threshold and $c^{(h,w)}_i$  represents the model's confidence scores in predicting the position $(h,w)$, as calculated by Eq. \ref{confidence}. These low-confidence pixels, which the teacher model itself struggles to distinguish, are likely to further mislead the student model. Although early S4 methods attempted to filter out unreliable pseudo-labels using confidence thresholds, simply discarding predictions below the threshold fails to address the problem, particularly in boundary regions. To address the issue of incorrect pseudo-labels, it is essential to explore a novel architecture that is independent of conventional S4 methods. Our VLM-PP is a novel module that operates independently of main S4 frameworks and can purify low-quality pseudo-labels as well as correct errors in them.

\subsection{VLM-PP Moudle}
\label{VLM-PP}
\begin{algorithm}[ht]
\caption{Training procedure of SemiEarth}
\label{alg1}
\begin{algorithmic}
\State 
\State $ \textbf{Input:} $
\State \hspace{0.5cm}$ D^U=\{(x_i^u)\}_{i=1}^{N_U} $, $D^L=\{(x_i^l,y_i)\}_{i=1}^{N_L} $
\State$\textbf{Output}$: 
\State\hspace{0.5cm}$\Theta$: optimal model parameters
\State 1:\textbf{while} until converge:
\State 2:\hspace{0.75cm}\textbf{for} $x_i^l$ , $x_i^u$  in  $ D^L$ , $ D^U$:
\State 3:\hspace{1.25cm}$\mathcal L_{S}$$=CE(model\_s(x_i^l),y_i)$
\State 4:\hspace{1.25cm}$x^{u,weak}_i=WeakAugment(x_i^u)$
\State 5:\hspace{1.25cm}$x^{u,strong}_i=StrongAugment(x_i^u)$

\State 6:\hspace{1.25cm}$p_i^{u,s}=model\_s(x^{u,strong}_i)$
\State 7:\hspace{1.25cm}$y^{u,t}=model\_t(x^{u,weak}_i)$
\State 8:\hspace{1.18cm}\textbf{for} pixels in $y^{u,t}$:
\State 9:\hspace{1.55cm}\textbf{if} pixels in $\mathcal{R}_{\text{low}}$:
\State 10:\hspace{1.75cm}Confidence estimation via VLM by Eq. \ref{generates_text}-\ref{VLM_confidence}

\State 11:\hspace{1.8cm}\textbf{if} pseudo-label matches VLM prediction:
\State 12:\hspace{2.05cm} Purify pseudo-labels via Eq. \ref{purify pseudo-labels}.
\State 13:\hspace{1.8cm}\textbf{else}: 
\State 14:\hspace{2.05cm} Rectify with VLM Pseudo-Labels.
\State 15:\hspace{1.15cm}\textbf{end for} 
\State 16:\hspace{1.15cm}$\mathcal L_{U}$$=CE(p_i^{u,s}, y^{u,t}_p)$
\State 17:\hspace{1.15cm}$\mathcal L=\mathcal L_{S}+\mathcal L_{U}$
\State 18:\hspace{1.15cm}Update $\Theta$ via gradient descent on $\mathcal L$
\State 19:\hspace{1.15cm}Save the best checkpoint $\Theta_{best}$
\State 20:\hspace{0.75cm}\textbf{end for}
\State 21:\textbf{return} $\Theta$
\State 22:\textbf{end}
\end{algorithmic}
\end{algorithm}

In this section, we introduce the scheme of VLM-PP for purifying pseudo-labels as briefly illustrated in Fig. \ref{VLM-PP_Fig}. VLM-PP leverages the powerful visual understanding capability of VLMs for pseudo-label verification and purification. In short, VLMs combine visual information with text prompts to generate predictions, which are then compared with the correct answers for consistency evaluation.

Specifically, given unsupervised data $x_i^u$ and a set of classes $\mathcal{N}={n_1,n_2,...,n_K}$,  we construct a classification prompt $\mathcal{P}$ that enforces the model to identify all visible semantic categories as follows.
\begin{equation}
\mathcal{P}=\text{\textit{List and locate all visible classes in the image.}}
\end{equation}  
The VLM model generates a text prediction through autoregressive decoding by combining prompts with visual information. The specific process is defined as follows:
\begin{equation}
w^u_i=\mathop{\arg\max}_w(\prod_{t=1}^T P_{\theta}(w_t|w_{<t},{x_i^u},\mathcal{P})),
\label{generates_text}
\end{equation}
where $\theta$ denotes the pretrained parameters of the generative VLM model, and $T$ is the length of the output sequence. The term $P_{\theta}(w_t|w_{<t},x_i^u,\mathcal{P})$ represents the conditional probability of generating a next word $w_{t}$ given an image $x_i^u$, prompt $\mathcal{P}$, and previously generated words $w_{<t}$. $w^u_i$ contains the category predicted by the VLM and the corresponding coordinates. To determine whether a candidate class is present in the generated sequence $w^u_i$ and to obtain the VLM's confidence score, we apply the following criterion.
\begin{equation}
c_i^{u,k} = \begin{cases}
\gamma & \text{, if class } n_k  \text{ is mentioned in } w^u_i, \\
0 & \text{, otherwise}.
\end{cases}
\end{equation}  
For class $n_k$, $c_i^{u,k}$ is the confidence score from the VLM, and $\gamma$ is a constant set to 0.95 in this paper. For categories with non-zero confidence scores, we refine their pixel-level alignments using the SAM model. Finally, softmax normalization is applied to produce the final VLM confidence $\tilde{c}_i^{{u,k}}$:
\begin{equation}
\tilde{c}_i^{{u,k}}= \frac{{c_i^{u,k}}}{{\sum}_{k=1}^{K}{c_i^{u,k}}+\epsilon
} \\,
\label{VLM_confidence}
\end{equation}
where $\epsilon$ is a small constant introduced for numerical stability to avoid division by zero. If the pseudo-label generated by the teacher model is of low confidence yet consistent with the VLM's predicted class, we perform pseudo-label purification as follows. 
\begin{equation}
\label{purify pseudo-labels}
\tilde{c}_i^{(h,w)} = 
\alpha_i^{(h,w)} \cdot c_i^{(h,w)} + (1-\alpha_i^{(h,w)}) \cdot \tilde{c}_{i}^{u,k(h,w)}.
\end{equation}
Here, $\alpha_i^{(h,w)} = \frac{c_i^{(h,w)}}{\tau_{\text{conf}}}$ is the purifying weight, $\tilde{c}_{i}^{u,k(h,w)}$ represents the VLM confidence score $\tilde{c}_i^{{u,k}}$ at the position $(h,w)$, and $\tilde{c}_i^{(h,w)}$ is the final pseudo-label confidence. In regions where the teacher model is highly confident, we directly adopt its pseudo-labels. In low-confidence regions, we perform pseudo-label purification to improve the reliability of unreliable pseudo-labels via adaptive fusion with VLM-derived knowledge. Moreover, for pixels with low confidence, the parameter $(1-\alpha_i^{(h,w)})$ in Eq.\ref{purify pseudo-labels} increases, i.e., strengthening the purifying weight of VLM. Even for pseudo-labels exhibiting extremely low confidence, VLM-PP can effectively purify them.

Additionally, when the teacher-generated pseudo-label conflicts with the VLM’s prediction, we replace it directly with the VLM’s output, as shown in Fig.\ref{VLM-PP_Fig}. We adopt the VLM’s confidence scores as the final pseudo-label confidence scores.
\begin{equation}
\label{rectify pseudo-labels}
\tilde{c}_i^{(h,w)} =  \tilde{c}_{i}^{u,k(h,w)}.
\end{equation}

Therefore, VLM-PP can rectify misclassified pseudo-labels produced by the teacher. Notably, any purified pseudo-labels that remain low-confidence are filtered out and excluded from the student model’s training.

Algorithm \ref{alg1} provides the core pseudo-code for the SemiEarth training process, enhancing the interpretability and clarity of its methodology. The core logic of VLM-PP is shown in lines 8 to 15. $D^U$ and $D^L$ are the unsupervised and supervised datasets for one epoch. $N_U$ and $N_L$ are number of images in $D^U$ and $D^L$. $model\_s $ and $model\_t$ denote the student and teacher models. $p_i^{u,s}$ denotes the student’s final prediction on unlabeled data. $y^{u,t}$ represents the pseudo-labels generated by the teacher. $\mathcal{R}_{\text{low}}$ denotes low-confidence region as defined in Eq. \ref{low-confidence regions}. $y^{u,t}_p$ denotes purified pseudo-labels. $CE$ is the Cross-Entropy loss function. 

Our RS S4 model is the first work to integrate Vision-Language Models (VLMs) into the RS S4 framework for pseudo-label purification. This section provides a thorough analysis of the motivation underlying the proposed VLM-PP model. Additionally, we elaborate on the core principles of the VLM-PP module, an independent component that not only enhances pseudo-label quality but also enables the correction of erroneous pseudo-labels. More importantly, SemiEarth departs from the conventional RS S4 architecture, offering strong interpretability while significantly outperforming state-of-the-art methods. It will be discussed in detail in the experimental section below.

\section{Experiments}
\label{Experiments}
In this section, we conduct experiments on RS datasets to validate our proposed novel RS S4 model. We first conduct quantitative and qualitative comparisons with SOTA models. Then, we perform ablation studies on SemiEarth to validate the effectiveness of the proposed modules. Our experimental code, dataset split files, and detailed experimental setup are all released at \href{https://github.com/wangshanwen001/SemiEarth}{https://github.com/wangshanwen001/SemiEarth}.

\subsection{RS Datasets and Data Augmentation}

\subsubsection*{\bf LoveDA}
The LoveDA RS dataset \cite{Loveda} consists of 5987 images containing a total of 166,768 annotated objects, collected from three distinct urban regions. Each image has a spatial resolution of 0.3 meters and a size of $1024\times1024$ pixels, and is labeled with one of seven semantic classes: building, road, water, barren, forest, agriculture, and background. Due to GPU memory limitations, all images are resized and cropped to $512 \times 512$ pixels, yielding 16,764 training samples for deep learning. Following the protocol of the previous SOTA work on RS S4\cite{huang2023semi}, using a 6:2:2 split into training, validation, and test sets for local evaluation, without submission to the dataset's competition website. The data we ultimately used to validate the model's performance is completely isolated from the training data. The specific processed datasets and data splits can be found in our open-source repository.

\subsubsection*{\bf ISPRS-Potsdam}
The ISPRS Potsdam RS dataset \cite{ISPRS_Potsdam} is widely used to advance research in semantic segmentation of RS imagery. It has a resolution of 0.05 meters and consists of 38 super-large satellite RS images, each with a size of $6000\times6000$ pixels. The dataset contains six segmentable classes, i.e., impervious surfaces, building, low vegetation, tree, car, and background. We crop each original image into $512 \times 512$ patches, resulting in a total of 5,472 cropped images for deep training. Consistent with the LoveDA dataset, we adopted the same data processing approach as the previous RS S4 SOTA, splitting the data into training, validation, and test sets in a 6:2:2 ratio. The data used to validate the model's performance is completely isolated from the training data.

\begin{figure}[!t]
\centering
\includegraphics[width=3.3in]{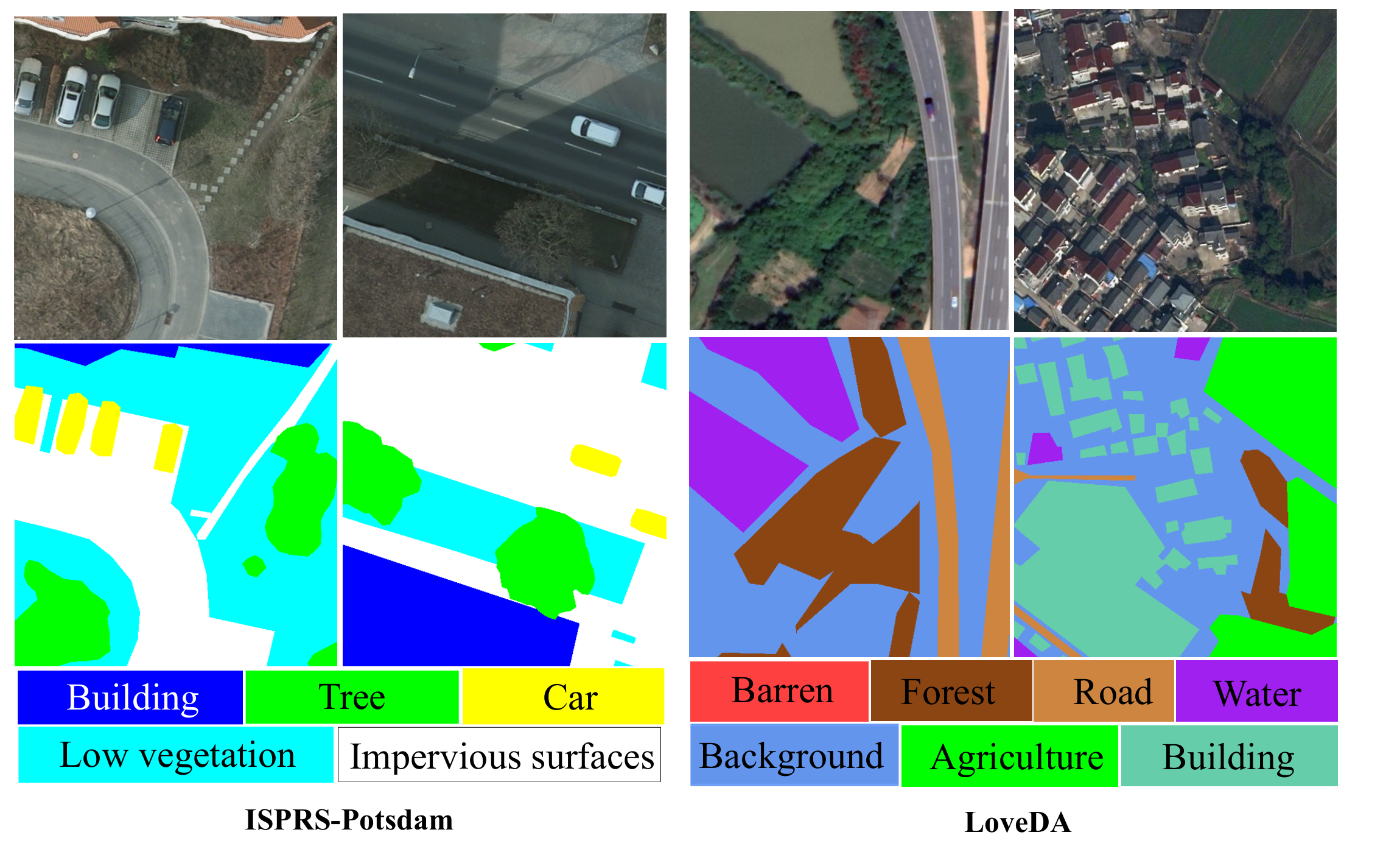}
\vspace{-0.35cm}
\caption{Samples from the two RS Datasets.}
\label{datasets}
\end{figure}

Some samples from these RS datasets are shown in Fig. \ref{datasets}, where categories are numerous, and boundaries among multiple classes are complex. We employed both weak and strong augmentations on the datasets. On unlabeled images, weak augmentation was performed with geometric transformations such as image scaling, horizontal flipping, and vertical flipping, while strong augmentation was implemented using methods such as photometric transformations, Gaussian blur, and CutMix \cite{yun2019cutmix}.

\begin{table}[ht]
\caption{Comparison results with SOTA methods on ISPRS-Potsdam dataset.\ The best results are highlighted in bold. IoU and mIoU are represented as percentages.\label{tab:table5}}
\centering
\normalsize
\scriptsize         
\setlength{\tabcolsep}{2pt} 
\begin{tabular*}{\linewidth}{@{\extracolsep{\fill}} c c c c c @{\hspace{1.25em}} c c c }
		\toprule
		\textbf{Ratio} & \textbf{Model} &\multicolumn{5}{c}{\textbf{IoU}}  & \textbf{mIoU} \\
		\cline{3-7}
		~ & ~ & Building & \makecell{Low \\ vegetation}  & Tree & Car & \makecell{Impervious \\ surfaces} &  ~\\
		\midrule
		\multirow{10}{*}{\textbf{1\%}}&  Mean teacher\cite{tarvainen2017mean} & 72.53 & 58.98 
         &60.43&67.97&67.39&65.46 \\
        ~& CutMix\cite{yun2019cutmix} & 55.58 & 42.05 &49.72&50.86&39.40&47.52\\
		~& CCT\cite{ouali2020semi} & 54.48 & 61.28 &48.56&52.95&60.71&55.59 \\
        ~& CPS\cite{chen2021semi} & 59.35 & 69.16 &62.89&59.88&66.33&63.52 \\
		~& LSST\cite{lu2022simple} & 68.74 & 75.24 &54.74&62.09&68.80&65.92 \\
		~& FixMatch\cite{sohn2020fixmatch} & 76.95 & 71.59 &64.71&65.85&72.81&70.38\\
		~& UniMatch\cite{yang2023revisiting} & 76.52 & 70.99 &65.44&66.62&72.64&70.44\\
        ~& DWL\cite{huang2024decouple} & 72.34 & \textbf{77.08}&62.74&62.57&72.22&69.39 \\
		~& AllSpark\cite{Wang_2024_CVPR} & 83.70 & 65.92 &59.64&69.77&75.31&70.87 \\
         ~& MUCA\cite{11062866} & 84.56 & 66.98 &56.96&71.52 & 76.64 & 71.33\\
		~& \textbf{Our (SemiEarth)} & \textbf{86.80} & 71.22 &  \textbf{71.96} &  \textbf{76.11} & \textbf{79.01}  & \textbf{77.02}  \\
		\midrule
		\multirow{10}{*}{\textbf{5\%}}& Mean teacher\cite{tarvainen2017mean} & 82.15 & 65.92 
         &67.11&72.21&74.60&72.40\\
		~& CutMix\cite{yun2019cutmix} & 52.94 & 68.86 &41.51&58.33&54.82&55.29 \\
		~& CCT\cite{ouali2020semi} & 72.90 & 80.25 &64.23&58.32&74.42&70.02\\
        ~& CPS\cite{chen2021semi} & 76.53 & 84.34 &57.98&69.45&75.39&72.74 \\
		~& LSST\cite{lu2022simple} & 69.26 & 84.55 &67.33&67.49&73.86&72.50 \\
        ~& FixMatch\cite{sohn2020fixmatch} & 78.12 & 74.87 &68.89&66.58&75.30&72.75\\
		~& UniMatch\cite{yang2023revisiting} & 78.24 & 73.59 &67.17&66.64&75.07&72.14 \\
        ~& DWL\cite{huang2024decouple} & 74.81 & \textbf{85.64} &66.38&62.99&75.68&73.10 \\
		~& AllSpark\cite{Wang_2024_CVPR} & 85.57 & 67.62 &60.61&73.48&77.15&72.88 \\
		~& MUCA\cite{11062866} & 88.45 & 69.53 &61.39&74.18&79.56& 74.62 \\
        	~& \textbf{Our (SemiEarth)} & \textbf{88.51} & 74.45 &  \textbf{74.06} &  \textbf{78.14} &  \textbf{79.87} & \textbf{79.01}  \\
		\midrule
		\multirow{10}{*}{\textbf{10\%}}& Mean teacher\cite{tarvainen2017mean} & 84.76 & 69.28 
         &68.83&71.66&76.51&74.21 \\
		~& CutMix\cite{yun2019cutmix} & 64.55 & 80.99 &64.79&65.50&68.01&68.77 \\
		~& CCT\cite{ouali2020semi} & 73.09 & 83.94 &61.12&60.45&73.06&70.33 \\
        ~& CPS\cite{chen2021semi} & 77.80 & 87.15 &61.12&68.48&75.89&74.09 \\
		~& LSST\cite{lu2022simple} & 70.92 & 86.06 &68.91&70.22&74.89&74.20 \\
        ~& FixMatch\cite{sohn2020fixmatch} & 77.97 & 76.17 &70.09&70.97&76.14&74.27 \\
		~& UniMatch\cite{yang2023revisiting} & 77.34 & 87.75 &70.79&56.65&76.46&73.80 \\
        ~& DWL\cite{huang2024decouple} & 76.37 & \textbf{88.42} &66.54&64.37&77.14&74.57 \\
		~& AllSpark\cite{Wang_2024_CVPR} & 86.29 & 69.83 &64.17&75.23&78.31&74.76 \\
		~& MUCA\cite{11062866} & 88.02 & 70.58 &64.53&75.20&79.92&75.65\\
        	~& \textbf{Our (SemiEarth)} & \textbf{90.59} & 75.44 & \textbf{75.01} & \textbf{79.64} & \textbf{83.24} & \textbf{80.78}\\
		\bottomrule
	\end{tabular*}
     \label{tab_Potsdam}
\end{table}

\begin{table}[!t]
\caption{Comparison results with other SOTA methods on the LoveDA dataset. The best results are in bold. IoU and mIoU are represented as percentages. \label{tab:table6}}
\centering
\scriptsize    
\setlength{\tabcolsep}{2pt} 
\begin{tabular*}{\linewidth}{@{\extracolsep{\fill}} c c c c c c c c c c}
		\toprule
		\textbf{Ratio} & \textbf{Model} &\multicolumn{7}{c}{\textbf{IoU}}  & \textbf{mIoU} \\
		\cline{3-9}
		~ & ~ & Bac &  Building &Road & Water & Barren & Forest & Agr &  ~\\
		\midrule
	\multirow{10}{*}{\textbf{1\%}}& Mean teacher\cite{tarvainen2017mean} & 44.73 & 42.53 &40.34&50.92&11.37&26.88&54.40&38.73\\
		~& CutMix\cite{yun2019cutmix} & 36.04 & 24.69 &10.03&24.60&3.43&6.67&10.19&16.52 \\
		~& CCT\cite{ouali2020semi} & 37.16 & 22.41 &27.86&43.98&14.51&25.38&36.67&29.71\\
        ~& CPS\cite{chen2021semi} & 46.52 & 20.87 &27.85&50.55&0.01&33.16&34.60&30.51 \\
		~& LSST\cite{lu2022simple} & 44.73 & 41.90 &39.90&62.65&29.27&31.26&48.29&42.57\\
        ~& FixMatch\cite{sohn2020fixmatch} & 46.78 & 51.20 &50.21&67.27&11.53&\textbf{36.79}&50.26&44.86 \\
		~& UniMatch\cite{yang2023revisiting} & 46.53 & 51.38 &49.36&\textbf{67.74}&10.86&33.40&52.28&44.51 \\
		~& DWL\cite{huang2024decouple} & 48.74 & 56.79 &51.59&63.42&22.56&35.20&55.38&47.67 \\
        ~& AllSpark\cite{Wang_2024_CVPR} & 63.87 & 47.70 &46.05&61.52&35.31&30.94&55.64&48.72\\
		~& MUCA\cite{11062866} & 64.89 & 56.03&47.14&63.86& 35.81&22.57& 58.18 &49.78 \\
         ~& \textbf{Our (SemiEarth)} & \textbf{64.99}  & \textbf{57.12}  & \textbf{56.89} & 56.86 & \textbf{73.17} &  25.69 & \textbf{59.03} & \textbf{56.25} \\
		\midrule
	\multirow{10}{*}{\textbf{5\%}} & Mean teacher\cite{tarvainen2017mean} & 49.73 & 46.22 &42.34&60.93&31.51&35.79&44.22&44.39 \\
		~& CutMix\cite{yun2019cutmix} & 41.48 & 41.62 &38.77&47.44&14.69&28.09&31.05&34.73 \\
		~& CCT\cite{ouali2020semi} & 46.80 & 44.62 &46.80&60.95&24.83&29.03&44.30&42.48\\
        ~& CPS\cite{chen2021semi} & 48.90 & 49.64 &47.97&60.27&4.67&36.09&47.32&42.12 \\
		~& LSST\cite{lu2022simple} & 51.48 & 45.66 &52.66&67.63&33.52&35.80&48.60&47.91 \\
        ~& FixMatch\cite{sohn2020fixmatch} & 45.40 & 53.05 &51.22&66.73&28.53&27.25&54.30&44.64 \\
		~& UniMatch\cite{yang2023revisiting} & 50.20 & 54.49 &50.46&67.18&26.79&30.06&54.86&47.72 \\
        ~& DWL\cite{huang2024decouple} & 48.75 & 55.00 &51.53&\textbf{69.49}&29.46&36.59&52.11&48.99 \\
		~& AllSpark\cite{Wang_2024_CVPR} & 65.09 & 55.06 &47.59&67.10&34.67&26.86&51.87&49.75\\
		~& MUCA\cite{11062866} & 67.29 & 56.04 &48.37&61.02&36.21&30.76& 57.09 & 50.97 \\
           ~& \textbf{Our (SemiEarth)} &  \textbf{69.57} & \textbf{57.70}  & \textbf{62.12} &  60.52 &\textbf{75.61} & \textbf{42.88} & \textbf{60.98} & \textbf{61.34} \\
		\midrule
		\multirow{10}{*}{\textbf{10\%}} & Mean teacher\cite{tarvainen2017mean} & 50.45 & 55.75 &43.56&66.15&35.24&36.96&45.64&47.68 \\
		~& CutMix\cite{yun2019cutmix} & 46.73 & 49.60 &47.36&59.99&29.06&37.77&40.60&44.44\\
		~& CCT\cite{ouali2020semi} & 44.07 & 45.22 &47.65&57.12&24.41&32.50&45.07&42.29 \\
        ~& CPS\cite{chen2021semi} & 51.30 & 54.93 &52.57&53.37&18.39&37.59&53.24&45.91 \\
		~& LSST\cite{lu2022simple} & 50.69 & 49.50 &52.63& 69.85 &27.25&36.24&52.06&48.32 \\
        ~& FixMatch\cite{sohn2020fixmatch} & 52.02 &  55.59 &53.20& 57.91& 25.86&40.83& 57.50& 48.99\\
		~& UniMatch\cite{yang2023revisiting} & 51.80 & 53.95 & 51.17&58.15&25.60&38.72&54.86&47.75 \\
        ~& DWL\cite{huang2024decouple} & 49.94 &  56.66 & 53.89& \textbf{70.35} & 30.62&41.49&53.13&50.87 \\
		~& AllSpark\cite{Wang_2024_CVPR} & 67.13 & 56.16 &40.67&63.58&32.54&32.03&56.91&49.86 \\
		~& MUCA\cite{11062866} & 68.69 & 58.20&41.82&65.62&37.09&35.01&57.38&51.97\\
          ~& \textbf{Our (SemiEarth)} & \textbf{71.10} & \textbf{60.24}& \textbf{64.87} & 62.44& \textbf{76.70} & \textbf{42.42}&\textbf{63.05}&\textbf{62.97}\\
		\bottomrule
	\end{tabular*}
    \label{tab_LoveDA}
\end{table}

\subsection{Evaluation Metric and Experimental Setup}

Following the standard evaluation protocol of previous RS S4 methods, we use mean Intersection-over-Union (mIoU) as the primary metric to assess model performance. The mIoU is calculated as the average IoU across all classes.
\begin{equation}
IoU_k = \frac{TP_k}{TP_k+FP_k+FN_k},
\end{equation}
\begin{equation}
mIoU=\frac{1}{K}\sum_{k=1}^{K} IoU_k,
\end{equation}
where $TP_k$, $FP_k$, and $FN_k$ represent true positives, false positives, and false negatives for class $k$, and $K$ is the total number of classes. ‌

Our model is implemented in PyTorch and trained on eight NVIDIA RTX 4090 GPUs. We adopt DINOv2\_small \cite{oquab2023dinov2} as the teacher-student backbone network and utilize Qwen-VL \cite{yang2025qwen2, yang2025qwen3} as the VLM. SemiEarth is trained for 50 and 20 epochs on the Potsdam and LoveDA datasets, respectively. Notably, previous RS S4 models also exclude the Clutter class in Potsdam and the Ignore class in LoveDA when  calculating the final mIoU \cite{xin2024confidence, lu2022simple, huang2024decouple,11062866}. To ensure consistency with prior work, we adopt the same evaluation protocol. Further implementation and experimental details can be found in our open-source repository.

\subsection{Quantitative Results compared to SOTA}

This section conducts experiments on ISPRS-Potsdam and LoveDA datasets, compared to the SOTA methods, including Mean teacher \cite{tarvainen2017mean}, CutMix \cite{yun2019cutmix}, CCT \cite{ouali2020semi}, CPS \cite{chen2021semi}, LSST \cite{lu2022simple}, FixMatch \cite{sohn2020fixmatch}, UniMatch \cite{yang2023revisiting}, DWL \cite{huang2024decouple}, Allspark \cite{Wang_2024_CVPR}, and MUCA\cite{11062866}. Specifically, we show the results for labeled data ratios of $1\%$, $5\%$, and $10\%$ to verify the effectiveness, respectively. Notably, for the previous SOTA methods, we adopt the network configurations as default in their papers and code, with some records directly referenced from their original papers.

The experimental results on the ISPRS-Potsdam and LoveDA datasets are shown in Table \ref{tab_Potsdam} and Table \ref{tab_LoveDA}, respectively. In Table\ref{tab_LoveDA}, "Bac" and "Agr" are abbreviations for the background and agriculture classes, respectively. We can observe that S4 methods, such as Mean teacher and CPS, achieve the poorest results. In contrast, S4 models like UniMatch and AllSpark yield relatively good results. This improvement attributes to their careful filtering or weighting of pseudo-labels in the unlabeled data, which reduces the harmful impact of low-confidence samples. DWL and MUCA devise domain-specialized architectures that significantly boost performance by addressing the unique characteristics of RS data, such as rich multiscale information and high interclass similarities. From the results, we can see that the proposed SemiEarth achieves the best mean mIoU across all classes on both RS datasets. 

In addition, SemiEarth not only achieves SOTA performance in quantitative comparisons with existing models but also marks a substantial step forward for RS S4 methods. Specifically, on the ISPRS-Potsdam dataset, with labeled data ratios of 1\%, 5\%, and 10\%, our model outperforms the second-best model MUCA by 5.69\%, 4.39\%, and 5.13\% in mIoU, respectively. Similarly, on the LoveDA dataset under 1\%, 5\%, and 10\% labeling ratios, our model dramatically outperforms all existing methods, surpassing the second-best model by substantial margins of 6.47\%, 10.37\%, and 11.0\% in mIoU, respectively.‌ This is because, although earlier S4 methods filtered out low-confidence pseudo-labels, such regions in RS images often correspond to class boundaries with high semantic ambiguity. Excluding pixels from these low-confidence regions during training compromises model performance. In contrast, SemiEarth introduces a novel purification strategy for low-confidence pixels, effectively addressing this limitation.

\subsection{Qualitative Results compared to SOTA}
\begin{figure*}[!t]
\centering
\includegraphics[width=7in]{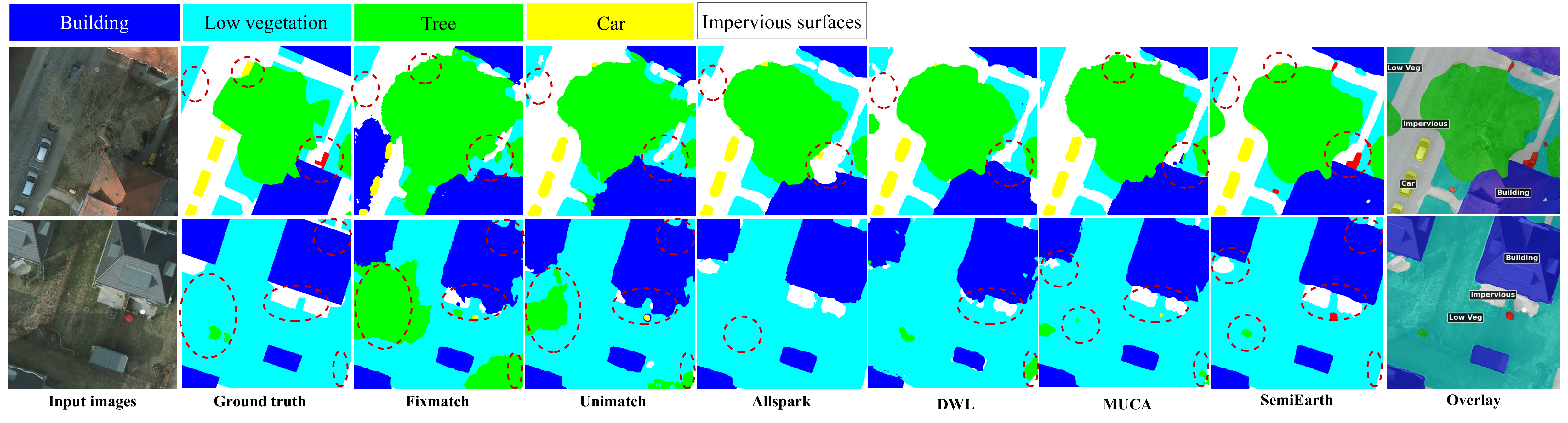}
\caption{Qualitative results with different SOTA S4 methods on the ISPRS-Potsdam dataset.}
\label{Potsdam}
\end{figure*}

\begin{figure*}[!t]
\centering
\includegraphics[width=7in]{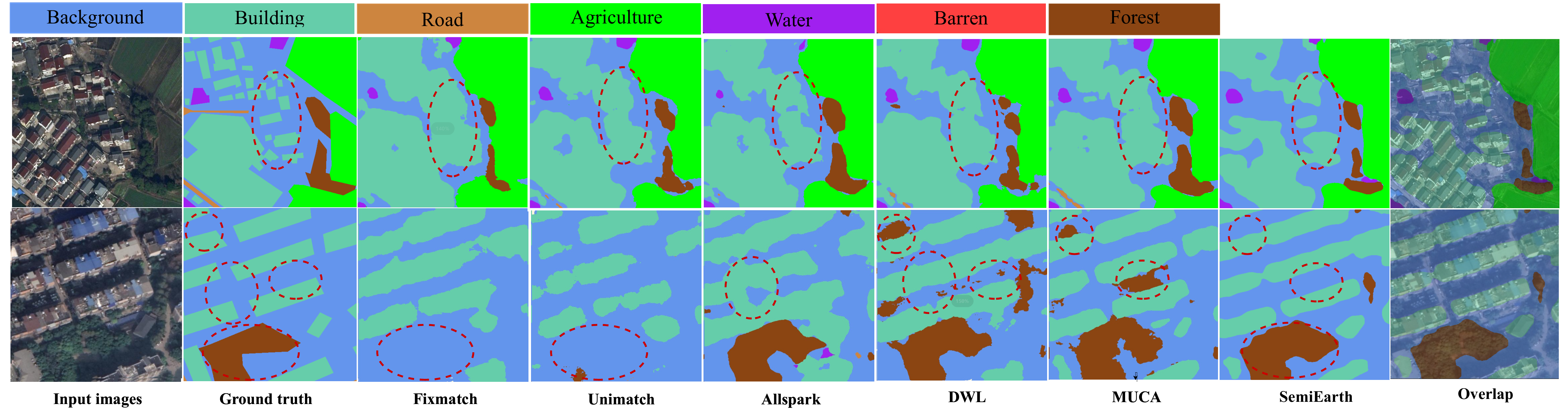}
\caption{Qualitative results with different SOTA S4 methods on the LoveDA dataset.}
\label{LoveDA}
\end{figure*}

To better illustrate the advantages of SemiEarth over other SOTA models, we provide a visual comparison of semantic segmentation results on RS datasets. Fig. \ref{Potsdam} and Fig. \ref{LoveDA} present qualitative comparisons with SOTA methods on the Potsdam and LoveDA datasets, respectively. For clarity, we highlight regions prone to mis-segmentation with dashed ellipses, emphasizing the comparatively accurate segmentation achieved by our model.

\vspace{-0cm}

\vspace{-0cm}

On the ISPRS Potsdam dataset, traditional S4 methods exhibit numerous misclassifications, particularly in complex scenes with complex and overlapping categories. Fixmatch and Unimatch make errors in segmenting regions for the Low vegetation, Tree, Impervious surfaces, and Building classes. The RS S4 models, such as DWL and MUCA, achieve promising performance in these regions, but they performed poorly at the boundaries of multiple categories. In comparison, SemiEarth achieves the highest segmentation accuracy among all evaluated methods. Additionally, we overlay our segmentation results on the original images and provide corresponding visual labels in the last column to facilitate comparison. The overlays demonstrate that our model produces highly accurate predictions, particularly along boundaries between multiple semantic classes. Notably, the small red region in the first row of the ground truth corresponds to the Clutter class defined in the official ISPRS-Potsdam dataset. However, due to its minimal presence in the scene, prior RS S4 methods excluded this class from quantitative evaluation \cite{huang2024decouple,11062866,li2021semisupervised} and also failed to segment it accurately in qualitative results. In contrast, SemiEarth correctly segments this challenging region.

On the LoveDA dataset, all the compared methods exhibit degraded visual quality. In particular, complex regions with intermixed classes suffer from segmentation errors in most approaches. Moreover, for categories requiring fine-grained fine-grained segmentation, such as the boundaries of isolated buildings and forest areas (see Fig. \ref{LoveDA}), most S4 models fail to produce accurate segmentation. In contrast, SemiEarth produces more accurate segmentations for classes including Building, Forest, Agriculture, and Background. To facilitate visual comparison, we overlay our segmentation results on the original images and provide class labels in the last column. These overlays demonstrate that our model yields higher-quality results on the LoveDA dataset, with improved boundary segmentation and fewer misclassifications compared to existing approaches.

\subsection{Ablation Study}

In this subsection, we conduct detailed ablation experiments on SemiEarth to validate the rationality of our model.

\subsubsection{Ablation Study of VLM-PP}
We perform ablation studies on the core component VLM-PP of SemiEarth model, i.e., \textit{without} and \textit{with} VLM-PP. In this subsection and the following ones, both the teacher and student networks employ DINOv2\_small as the backbone and Qwen-VL as VLM, with a labeled data ratio of 5\%.

\begin{table}[!t]
\caption{Ablation of the VLM-PP. \label{Ablation of VLM-PP}}
\centering
\footnotesize
\begin{tabular}{ c |c |c}
		\toprule
		\textbf{Dataset} & \textbf{Network} &\textbf{mIoU} \\
		\midrule
		 \multirow{3}{*}{ISPRS-Potsdam} & Without VLM-PP  & 74.42  \\
        \cmidrule{2-3}
        ~ &  VLM-PP &   79.01 \\
         \midrule
          \multirow{3}{*}{LoveDA} & Without VLM-PP   & 56.33  \\
        \cmidrule{2-3}
         ~ &  VLM-PP & 61.34  \\
		\bottomrule
	\end{tabular}
\end{table}

The experimental results are shown in Table \ref{Ablation of VLM-PP}. It can be seen that without the proposed core component VLM-PP, the performance is low. Meanwhile, the performance improves significantly with VLM-PP. Specifically, the mIoUs are increased by 4.59\% and 5.01\% on the Potsdam and LoveDA RS datasets, respectively. The experimental results demonstrate that our proposed VLM-PP significantly enhances the performance of the S4 model in RS domain.

\begin{figure}[h]
\centering
\includegraphics[width=3.5in]{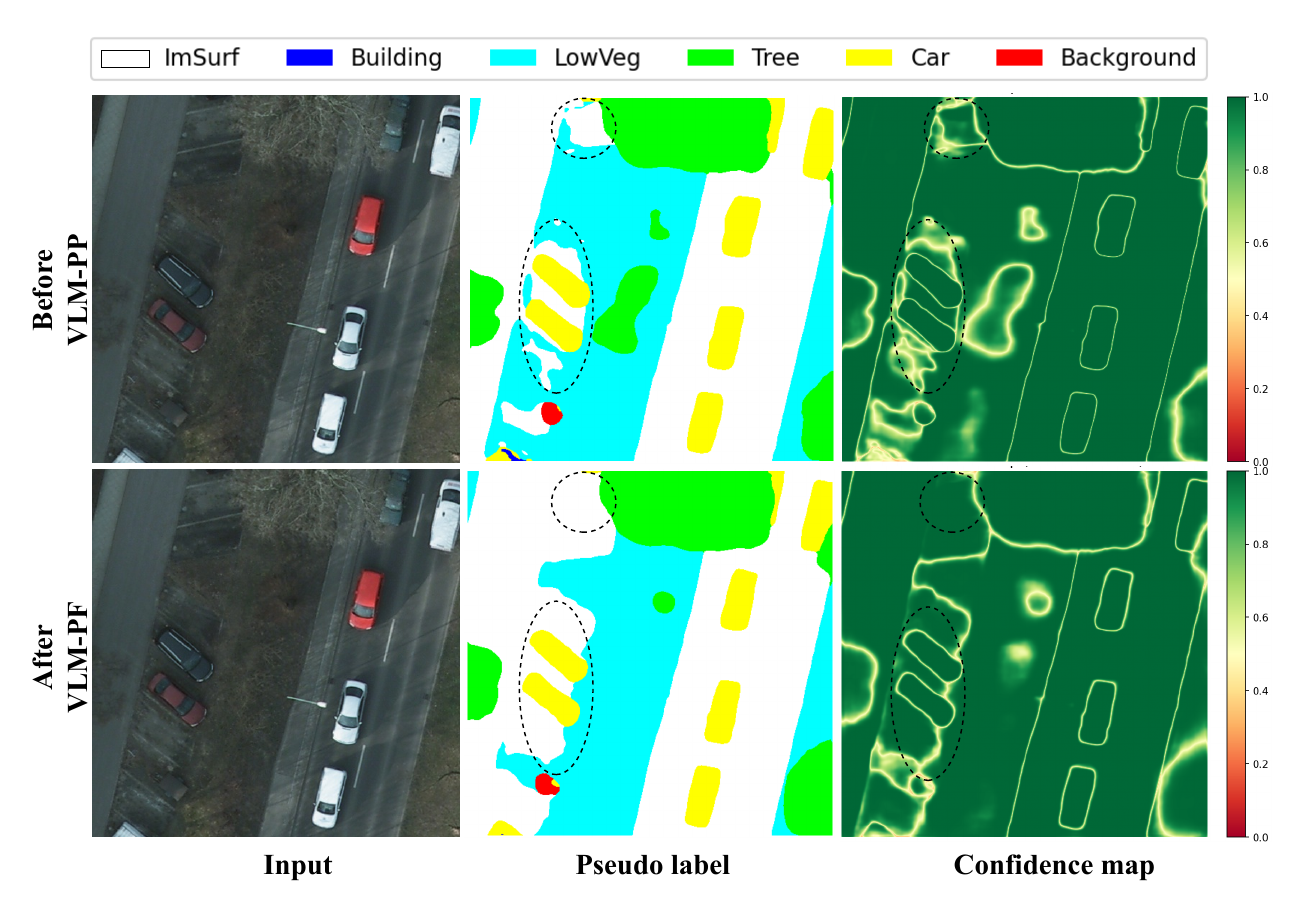}
\caption{Before and after the VLM-PP module: comparison of the quality of pseudo-labels and confidence.}
\label{pseudo}
\end{figure}
\vspace{0cm}
\subsubsection{Visualized Results of the VLM-PP Module}
\label{Visualized VLM-PP}
To better illustrate the effectiveness of the VLM-PP module, we visualize the pseudo-labels and their confidence scores generated by the teacher model—both before and after applying VLM-PP within the same training iteration.

\begin{figure}[!t]
\centering
\includegraphics[width=3.3in]{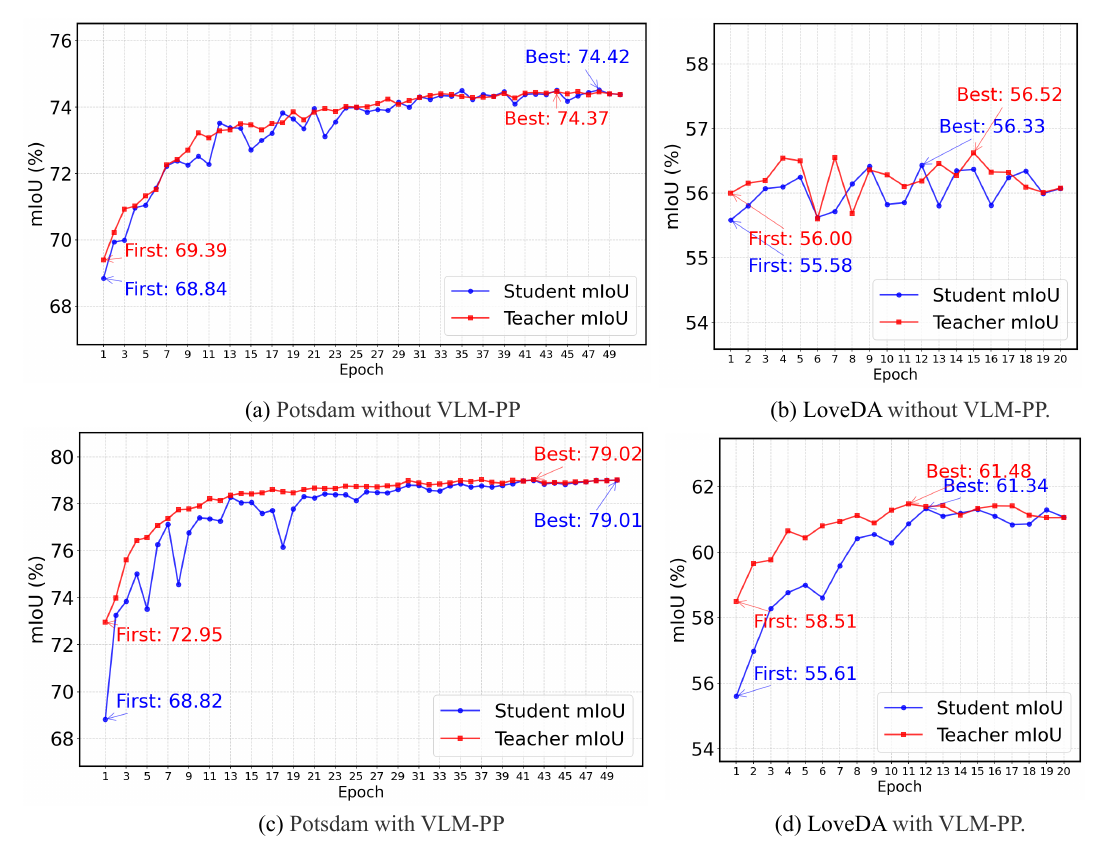}
\caption{VLM-PP significantly improves the quality of pseudo-labels generated by the teacher model, enabling it to effectively guide the student model during training iterations.}
\label{VLM-PP_Iteration}
\end{figure}
\vspace{0cm}

The results are shown in Fig. \ref{pseudo}. In complex regions of RS images, particularly at boundaries among Low vegetation, Car, Tree, and Impervious surfaces, the pseudo-labels generated by the teacher network before VLM-PP are spatially fragmented. Moreover, their associated confidence scores are consistently low, reflecting the teacher's uncertainty in assigning labels to these multi-class boundary zones. Such unreliable supervision can mislead the student model into learning erroneous patterns. In contrast, after applying the VLM-PP module, the teacher produces pseudo-labels with significantly higher confidence and improved semantic consistency in these challenging boundary zones. This is because VLM-PP not only reinforces correctly predicted regions by boosting their confidence but also rectifies misclassified pixels, thereby yielding more accurate and reliable pseudo-labels for student training.

\subsubsection{Purified Pseudo-Labels Improve Student Learning}
In this section, we visualize how teacher network progressively guides the student network with purified pseudo-labels throughout the training process. The teacher model generates pseudo-labels that serve as supervision for the student, and we assess the quality of these pseudo-labels using the mIoU metric. Fig. \ref{VLM-PP_Iteration} compares the SemiEarth iteration dynamics under two settings: without and with VLM-PP purification. 

In the beginning of training, the performance of student is nearly identical with and without VLM-PP. Then, the quality of pseudo-labels from teacher becomes crucial for effective learning. Further analysis reveals that the segmentation accuracy of teacher without VLM-PP is only marginally better than (or even comparable to) that of the student  is. Therefore, the teacher offers little effective supervision to the student. Moreover, in the absence of VLM-PP, both models rapidly overfit the limited labeled data as training progresses. Their learning dynamics become tightly coupled because the parameters of teacher are updated via EMA from the student and lack a mechanism to refine low-quality pseudo-labels. Once the teacher can no longer generate pseudo-labels that are more accurate than the student's own predictions, their performance fluctuate in lockstep, showing no consistent improvement.

In contrast, with VLM-PP enabled, the teacher provides consistently high-quality supervision from the early stages of training. Its pseudo-labels outperform the student’s own predictions, enabling the student to progressively learn accurate semantic representations. This advantage stems from VLM-PP’s ability to purify the teacher’s pseudo-labels. 

\begin{figure}[!t]
\centering
\includegraphics[width=3.6in]{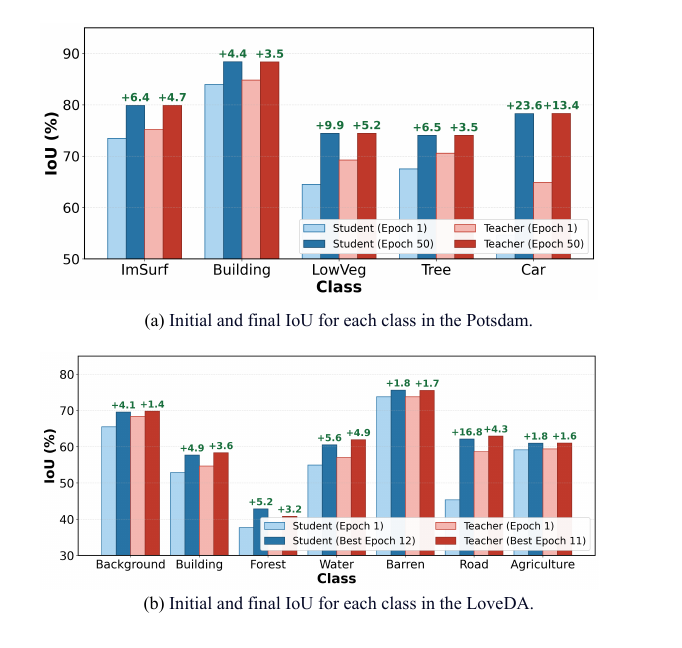}
\caption{The improvement in IoU from the start to the end of training for each class on the RS datasets.}
\label{IoU_class}
\end{figure}
\vspace{-0 cm}

Additionally, as illustrated in Fig. \ref{IoU_class}, we visually present the per-class IoU improvements of our model on the RS datasets over the course of training. On both datasets, the student model exhibits the most significant gains for the Car and Road classes, with IoU increases of 23.6\% and 16.8\%, respectively. These categories are particularly challenging, e.g., Car occupies only a small fraction of pixels in RS images, while Road has irregular and ambiguous boundaries. The strong performance on these difficult classes can be attributed to the high-quality pseudo-labels consistently generated by the teacher model throughout training.

\subsubsection{Purification Confidence Threshold in VLM-PP}
\begin{figure}[!t]
\centering
\includegraphics[width=3.4in]{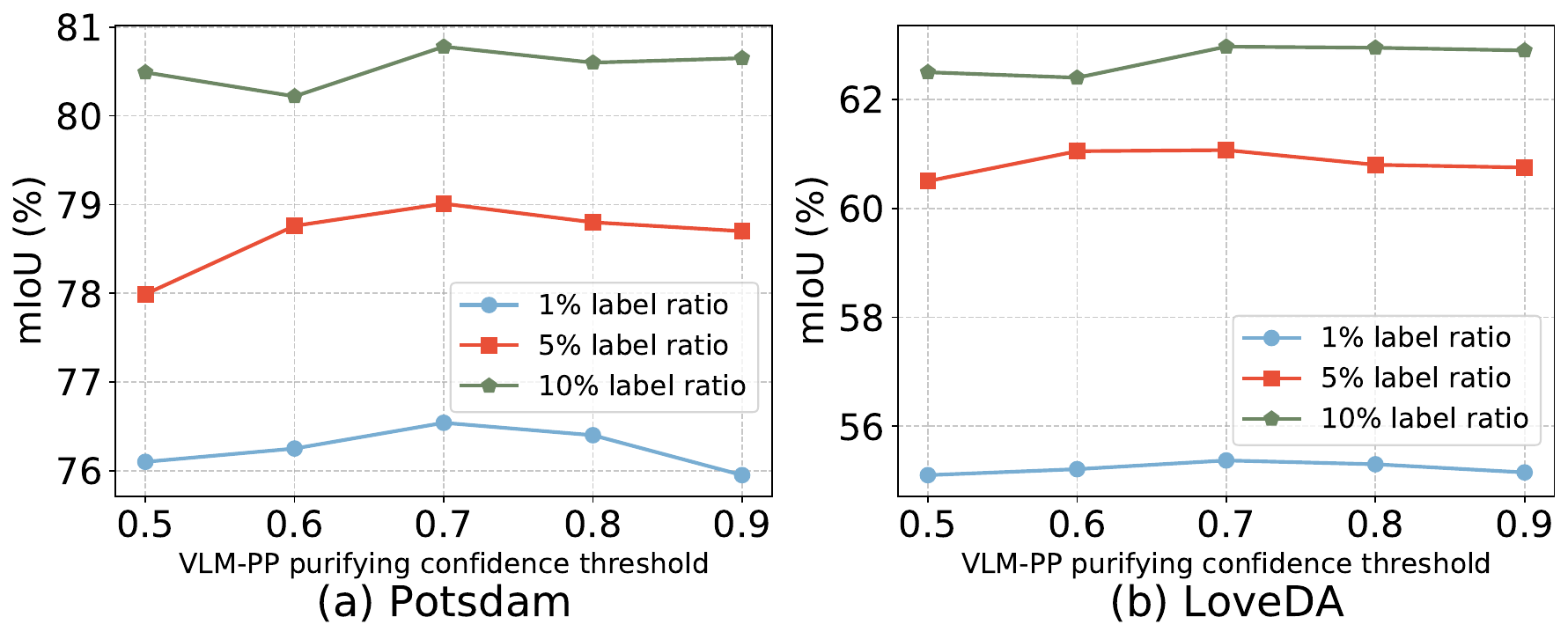}
\caption{Values of the different purifying confidence thresholds.}
\label{purifying_threshold}
\end{figure}

We investigate the optimal confidence threshold for VLM-PP to purify pseudo-labels generated by the teacher model. Specifically, we evaluate thresholds ranging from 0.5 to 0.9, with results presented in Fig. \ref{purifying_threshold}. As the purification threshold increases, the model's mIoU initially rises and then declines. This trend can be explained as follows: a threshold that is too low fails to filter out low-quality pseudo-labels, thereby introducing noise that misleads the student model; conversely, a threshold that is too high discards even highly confident predictions from the teacher, unnecessarily reducing the amount of useful supervision. The best performance is achieved within the range of 0.7–0.8, and we adopt 0.7 as the final threshold for all comparative experiments.

\vspace{-0cm}

This subsection validates the effectiveness and necessity of the proposed components through systematic ablation studies. The VLM-PP module significantly enhances performance on RS S4, while its underlying mechanism provides strong interpretability.

\section{Conclusion}
\label{Conclusion}
Our study addresses the challenge of low-quality pseudo-labels associated with semi-supervised RS image semantic segmentation by proposing the SemiEarth model. SemiEarth adopts an overall teacher-student model architecture and incorporates a specialized pseudo-label purification module named VLM-PP. VLM-PP, as a novel and independent module, aims to purify the low-quality pseudo-labels provided by the teacher model, preventing the student model from being misled. Compared to other state-of-the-art methods, our model achieves the highest mIoU on RS datasets and, unlike most prior works, provides strong interpretability. Specifically, detailed ablation studies demonstrate that VLM-PP consistently and reliably purifies low-quality pseudo-labels across all training iterations, thereby boosting the student model's learning performance. Nevertheless, current semi-supervised methods for remote sensing (RS) image analysis still encounter several key bottlenecks. For instance, while this paper pioneers the integration of Vision-Language Models into the RS S4 domain, the potential of combining large AI models with complex spectral RS imaging,. e.g., hyperspectral data, within the S4 framework remains largely unexplored. Our future work will focus on investigating how VLMs can be effectively leveraged for such challenging complex spectral RS imaging in S4. Finally, as the first work to introduce Vision-Language Models (VLMs) into the RS S4 domain, we hope that SemiEarth will serve as a valuable benchmark and help catalyze future research in this emerging direction.

\section*{Acknowledgments}
We would like to express our sincere appreciation to the anonymous reviewers.

\bibliographystyle{IEEEtran}
\bibliography{ref}

\vspace{11pt}
\vfill

\end{document}